\documentclass[10pt,twocolumn,letterpaper]{article} %
\usepackage{amsfonts,amssymb}
\usepackage{iccv}
\usepackage{stfloats}
\usepackage{caption}
\usepackage{times}
\usepackage{epsfig}
\usepackage{graphicx}
\usepackage{amsmath}
\usepackage{amssymb}
\usepackage{enumitem}
\usepackage[table,xcdraw]{xcolor} 
\usepackage{booktabs} 
\usepackage{cite}
\usepackage[marginal]{footmisc}
\usepackage[english]{babel}
\usepackage{appendix}
\usepackage{url}
\usepackage{tablefootnote}
\makeatletter
\newcommand{\spewnotes}{%
	\tfn@tablefootnoteprintout%
	\global\let\tfn@tablefootnoteprintout\relax%
	\gdef\tfn@fnt{?*?}%
}
\makeatother

\usepackage[pagebackref=true,breaklinks=true,letterpaper=true,colorlinks,bookmarks=false]{hyperref}

\iccvfinalcopy 

\begin{document}
\title{PI-REC: Progressive Image Reconstruction Network With Edge and Color Domain}
\renewcommand{\thefootnote}{\fnsymbol{footnote}}

\twocolumn[{ \author{Sheng You\\
		Nanjing University, China\\
		{\tt\small mf1832226@smail.nju.edu.cn}
		\and
		Ning You\\
		Sun Yat-sen University, China\\
		{\tt\small youn7@mail2.sysu.edu.cn}
		\and
		{Minxue Pan\tablefootnote{* Corresponding author}} \\ %
		Nanjing University, China\\
		{\tt\small mxp@nju.edu.cn}
	} \maketitle
	\centering
		\includegraphics[width=0.95\linewidth]{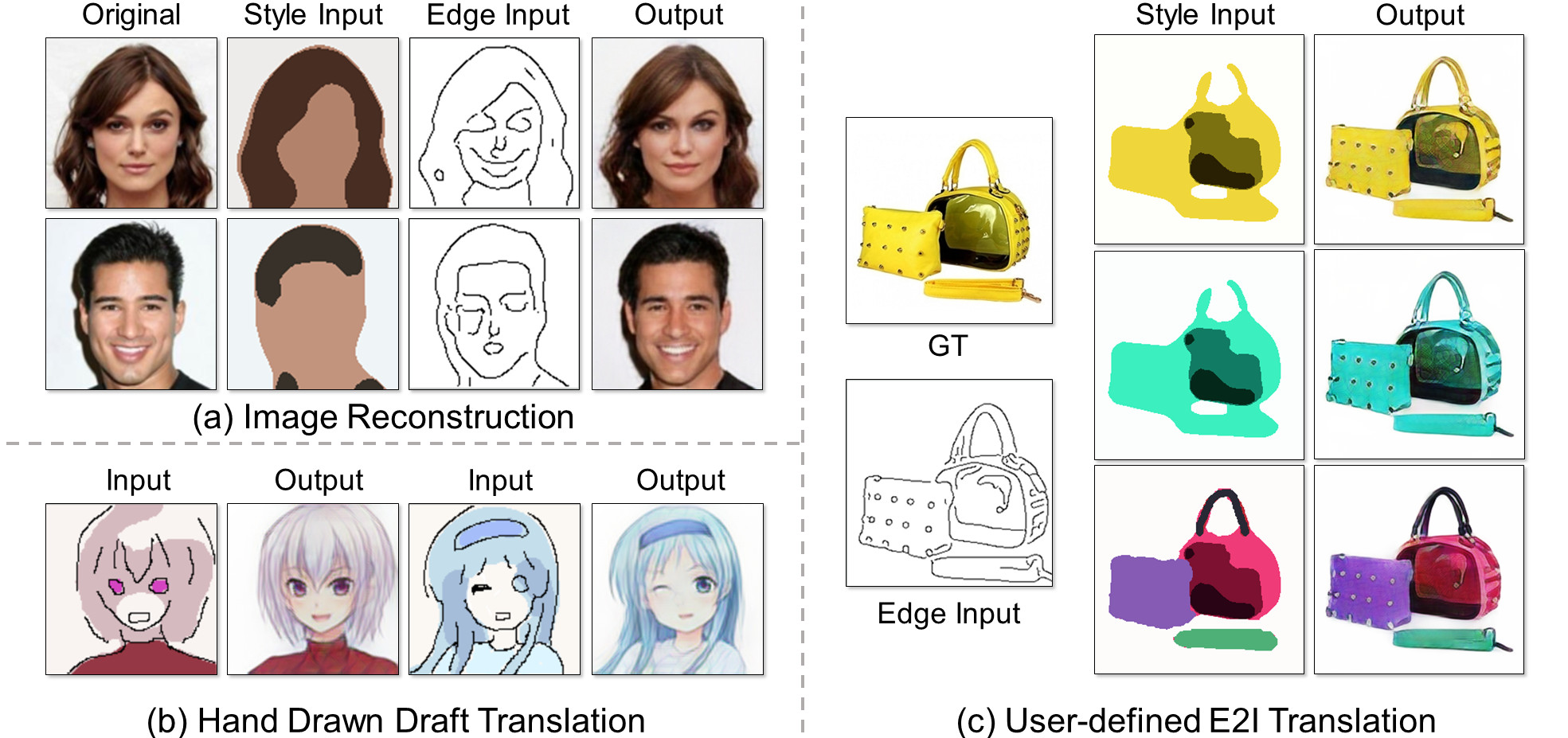}
		\captionof{figure}{(a) \textbf{Image reconstruction results.} Our method enables reconstructing lifelike images from extracted sparse edge and flat color domain. (b) \textbf{Hand drawn draft translation.} From draft drawn by hand, our method synthesizes accurate and refined images. (c) \textbf{User-defined E2I translation.} Users are allowed to obtain desired output accurately by feeding user-defined and pixel-level style images to our model.}
		\label{fig:banner}
\vskip 7mm
} ]
\spewnotes
\begin{abstract}
   We propose a universal image reconstruction method to represent detailed images purely from binary sparse edge and flat color domain. Inspired by the procedures of painting, our framework, based on generative adversarial network, consists of three phases: Imitation Phase aims at initializing networks, followed by Generating Phase to reconstruct preliminary images. Moreover, Refinement Phase is utilized to fine-tune preliminary images into final outputs with details. This framework allows our model ge\-ne\-rating abundant high frequency details from sparse input information. We also explore the defects of disentangling style latent space implicitly from images, and demonstrate that explicit color domain in our model performs better on controllability and interpretability. In our experiments, we achieve outstanding results on reconstructing realistic images and translating hand drawn drafts into satisfactory paintings. Besides, within the domain of edge-to-image translation, our model PI-REC outperforms existing state-of-the-art methods on evaluations of realism and accuracy, both quantitatively and qualitatively.
\end{abstract}

\begin{table*}[ht]
\begin{center}

\setlength{\tabcolsep}{2mm}{
\begin{tabular}{rcccccc}
\toprule
& \multicolumn{1}{c}{\begin{tabular}[c]{@{}c@{}}{SketchyGAN}\\ {\cite{chen2018sketchygan}}\end{tabular}} & \multicolumn{1}{c}{\begin{tabular}[c]{@{}c@{}}{Scribbler}\\ {\cite{sangkloy2017scribbler}}\end{tabular}} & \multicolumn{1}{c}{\begin{tabular}[c]{@{}c@{}}{Sparse Contour}\\ {\cite{dekel2018sparse}}\end{tabular}} & \multicolumn{1}{c}{\begin{tabular}[c]{@{}c@{}}{MUNIT}\\ {\cite{huang2018multimodal}}\end{tabular}} & \multicolumn{1}{c}{\begin{tabular}[c]{@{}c@{}}{BicycleGAN}\\ {\cite{zhu2017toward}}\end{tabular}} &   \multicolumn{1}{c}{\begin{tabular}[c]{@{}c@{}}{\bf PI-REC}\\ {\bf (ours)}\end{tabular}} \\ \midrule \midrule
Domain             & S2I    & S2I     & IR    & I2I    & I2I    & IR                                                                           \\
\rowcolor[HTML]{ECF4FF}
Sparse content$^{\dag}$      & \checkmark   & -   & \checkmark  & \checkmark   & \checkmark  & \checkmark                                                                           \\
Dense content$^{\dag}$     & -      & \checkmark   & \checkmark  & \checkmark   & \checkmark  & \checkmark                                                                              \\
\rowcolor[HTML]{ECF4FF}
Example-guided style$^{\dag}$     & -   & -   & -   & \checkmark     & \checkmark    & \checkmark                                                                            \\
User-defined style$^{\dag}$    & -   & \checkmark    & -    & -     & -    & \checkmark                                                                          \\
\rowcolor[HTML]{ECF4FF}
Hand drawn draft compatibility
 & \checkmark     & -    & -    & -     & -    & \checkmark                                                                           \\
\toprule[0.2pt]
High fidelity content$^{\ast}$ & -      & \checkmark   & \checkmark  & \checkmark   & \checkmark  & \checkmark
\\
\rowcolor[HTML]{ECF4FF}
High fidelity style$^{\ast}$
& -     & \checkmark    & \checkmark    & -     & -    & \checkmark                                                                           \\

\toprule
\end{tabular}}

\end{center}
\caption{ \textbf{Main dissimilarities among correlative major methods across domains of S2I synthesis, I2I translation and IR.} $^{\dag}$ denotes various features of inputs and $^{\ast}$ represents output quality.}
\label{table:models}
\end{table*}

\section{Introduction} \label{Introduction}

Image reconstruction (IR) is essential for imaging applications across the physical and life sciences, which aims to reconstruct the image from various information given by the ground truth one.

Generally, an image is the composition of content and style. Sketch extracted from image or drawn by hand is commonly used as content \cite{chen2018sketchygan,eitz2011photosketcher,chen2009sketch2photo} in the domain of sketch-to-image (S2I) synthesis. However, sketch that contains dense detailed information like line thickness and boundary intensity is hard to edit or draw. A binary contour map with gradients \cite{dekel2018sparse} can also be utilized to represent images, but only in the domain of image editing. In short, the content extracted by the abovementioned methods are not sparse and manageable enough.

Recently in the domain of image-to-image (I2I) translation\cite{isola2017image,huang2018multimodal,zhu2017toward}, one can synthesize photo-realistic images from sparse binary edge maps, employing a cycled framework based on conditional generative adversarial networks (cGANs) \cite{mirza2014conditional}. These methods disentangle the image in order to extract content and style respectively. However, in the field of edge-to-image (E2I) translation, the input of example-guided style cannot reconstruct high-fidelity style or color in output accurately.

These aforementioned limitations lead us into con\-si\-de\-ring how we can solve the conflicts between sparser inputs and more controllable style space. Our work here is partly motivated by the procedure during painting, the construction of which can be summarized into three parts: copy drawing, preliminary painting and fine-tuned piece. Many aspiring young artists are advised to learn by copying the masters at the beginning. During preliminary painting, sketching and background painting provide basic elements and structure information. At fine-tune stage, the piece are gradually refined with details, laying on increasingly intense layers of color, which add lights and shadow.

In analogy to such painting process, we propose a universal image reconstruction method to represent detailed images with binary sparse edge and flat color domain \cite{jo2019sc}. The inputs of binary edge and color domain are sparse and easy enough to be extracted (Figure \ref{fig:banner} (a)), to be hand drawn (Figure \ref{fig:banner} (b)) or to be edited (Figure \ref{fig:banner} (c)). We input the color domain as explicit style feature instead of extracting implicit latent style vector in I2I translation, in order to improve the controllability and interpretability on image styles. Our model based on generative adversarial network consists of three phases in turn: \emph{Imitation} \emph{Phase}, \emph{Generating} \emph{Phase} and \emph{Refinement} \emph{Phase}, which correspond to painting procedures, respectively. Within the domain of E2I translation our model PI-REC shows promi\-sing performance on the user-defined style tests from sparse input as shown in Figure \ref{fig:banner}. It can generate more accurate content details with color style than the former methods. Our code is available at \url{https://github.com/youyuge34/PI-REC/}.

Our key contributions can be summarized as:
\begin{itemize}
\item
We propose a novel universal image reconstruction architecture, where the progressive strategy used endows our model PI-REC with the ability of reconstructing high-fidelity images from sparse inputs.
\item
We improve the controllability and interpretability by using flat color domain as explicit style input instead of extracting latent style vector frequently used in I2I translation.
\item
We propose the hyperparameter confusion (HC) ope\-ra\-tion for PI-REC to achieve remarkable hand drawn draft translation results, in the hope of promoting the development of auto painting technology.
\end{itemize}

\begin{figure*}[t]
\centering


\includegraphics[width=1\linewidth]{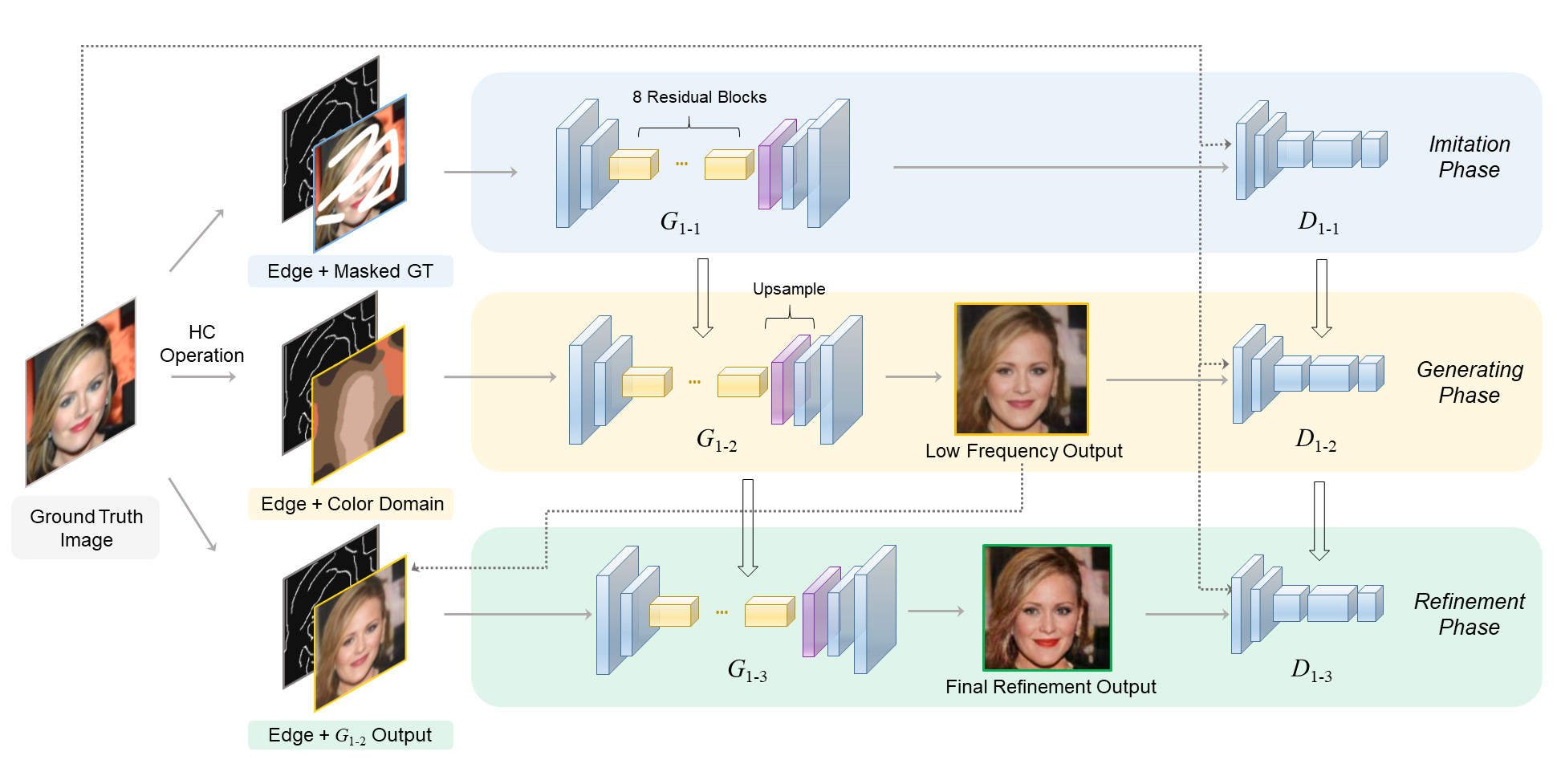}

\caption{\textbf{Network architecture of our proposed model PI-REC}. It contains three phases: \emph{Imitation} \emph{Phase}, \emph{Generating} \emph{Phase} and \emph{Refinement} \emph{Phase} with only one generator and one discriminator trained progressively.}
\label{Network}
\end{figure*}

\section{Related Work} \label{Related Work}
Image reconstruction (IR), as an interdisciplinary subject, has made great progress with the development of deep learning in recent years. Various information extracted from original images can be used to reconstruct the ground truth one. This idea is incorporated into massive fields including image editing, image inpainting and image translation. In our paper, we focus on reconstructing images from sparse inputs of content and style.

{\bf Generate adversarial network (GAN).} Definitely, GAN has been one of the most prevalent theories since the birth of the vanilla one \cite{goodfellow2014generative}. In the training phase of GAN, a generator is trained with a discriminator alternately with the intention of generating desired output. The basic idea of internal competition can be extended to image reconstruction, in order to generate realistic details.

{\bf Sketch-to-image (S2I) synthesis.} The main methods of S2I synthesis domain could be divided into two: indirect retrieval and direct synthesis. Sketch Based Image Retrieval (SBIR) attempts to bridge the domain gap between features extracted from sketches and photos \cite{cao2011edgel,cao2010mindfinder,eitz2010evaluation,eitz2011sketch}. However, bag-of-words models with lots of extracted features \cite{mao2019mode} are problematic to match edges with unaligned hand drawn sketches. Cross-modal retrieval is applied into S2I synthesis problem using deep neural networks, which is able to do instance-level \cite{yu2016sketch,sangkloy2016sketchy} or category-level \cite{srivastava2017veegan,che2016mode} S2I retrieval. Nevertheless, It is challenging for SBIR to complete pixel-level synthesis or to consider style as input owing to the self-limitation of retrieval. Scribbler \cite{sangkloy2017scribbler} succeeds to introduce GAN into S2I synthesis field without retrieval, which uses dense sketch and color stroke as inputs. However, color stroke as style input confuses the network about which
area to colorize when content input is sparse. SketchyGAN \cite{chen2018sketchygan} has a truly sparse sketch input while the style cannot be user-defined.

{\bf Image-to-image (I2I) translation.} Isola $et$ $al$. \cite{isola2017image} proposes the first unified framework Pix2Pix for I2I translation utilizing conditional GANs (cGANs) \cite{mirza2014conditional},using semantic label map or edge as input. It has an overall capability on diverse image translation tasks inc\-lu\-ding edge-to-image (E2I) translation. Based on these fin\-dings, CycleGAN \cite{zhu2017unpaired} introduced cycle-consistency loss and exploit cross-domain mapping for unsupervised training. However, the methods above are only appropriate to one-to-one domain translation. Recent researches focus on multi-modal I2I translation\cite{choi2018stargan,romero2018smit,anoosheh2018combogan} tasks which could transform images across domains. The random latent style is merged into the structure of pix2pixHD \cite{wang2018high} to generate diverse styles, which is still uncontrollable. BicycleGAN \cite{zhu2017toward} includes style vector bijection and self-cycle structure into the generator in order to output diverse reconstructions. However, its style of output from example-guided style image is not accurate under complex cases. We explore the defects further in Section \ref{Qualitative Evaluation}. Unsupervised multi-modal I2I translation methods \cite{lee2018diverse} are proposed to fit the unpaired datasets. Whereas, in our subject of reconstruction from sparse information, edges we need could be extracted from original images to form paired datasets. Thus, adopting unsupervised training in our research is redundant.

Table \ref{table:models} summarizes main dissimilarities regarding the literature for representative and correlative methods across domains of S2I synthesis, I2I translation and IR. PI-REC has more capabilities than prior methods in that it takes sparser edges and pixel-level color style as inputs to generate images with both high-fidelity in content and style.

\section{PI-REC} \label{PI-REC}The ultimate purpose of our work is to reconstruct lifelike image purely from binary sparse edge and color domain.  Thus, we propose PI-REC model architecture which consists of three phases in turn: \emph{Imitation} \emph{Phase}, \emph{Generating} \emph{Phase} and \emph{Refinement} \emph{Phase} with only one generator and one discriminator. During training, exploiting progressive strategy on the same generator reduces the time cost and RAM memory cost.

\subsection{Preprocessing of training Data} \label{Preprocessing of training Data}

{\bf Edge.}
Edges are treated as the content of an image in our method. We choose Canny algorithm \cite{canny1987computational} to get rough but solid binary edges instead of dense sketches extracted by HED \cite{xie2015holistically}, which enhances the generalization capability of our model with relatively sparser inputs.

{\bf Color domain.}
Color domain corresponding to the style features is extracted in an explicit way. We apply a median filter algorithm followed by K-means \cite{coates2012learning} algorithm to obtain the average color domain.  After that, we use a median filter again to blur the sharpness of the boundary lines.

{\bf Hyperparameter confusion (HC).}
When extracting edges or color domains from input images, there are several algorithms that require hyperparameters. During training, we adopt different random values of hyperparameters in a range, which can augment the training datasets to prevent overfitting. Not only that, each pixel in the extracted edge has a 8\% chance to be reset to value zero, on account of the diverse cases, where some people draw or edit casually while others paint elaborately. HC operation enhances generalization ability of our model to deal with the complex hand drawn draft translation cases, which is presented in Section \ref{Ablation Study}.

\subsection{Model Architecture} \label{Model Architecture}
As shown in Figure \ref{Network}, our progressive architecture is based on three phases: \emph{Imitation} \emph{Phase}, \emph{Generating} \emph{Phase} and \emph{Refinement} \emph{Phase}. We denote our generator and discriminator as $G_1$ and $D_1$ respectively. The details are described below.

{\bf Generator.}
$G_{1\text{-}1}$, $G_{1\text{-}2}$ and $G_{1\text{-}3}$ represent the three training phases of our generator $G_1$, each in due succession. Only when the network converges in the current phase, can our model enter into the next training phase. The architecture of $G_1$ is based on U-net \cite{ronneberger2015u} and  Johnson \emph{et} \emph{al}. \cite{johnson2016perceptual}. Specifically, $G_1$ network employs encoder and decoder structure with eight residual blocks \cite{he2016deep} merged into middle part, utilizing dilated convolutions in convolution layers. Since our method has three stages to optimize the image quality progressively, the redundant skip connections between layers of encoder and decoder are removed. In addition, \emph{checkerboard} \emph{artifact} is a serious problem \cite{odena2016deconvolution} occurring when deconvolution is used. To tackle the problem, we replace the first deconvolution layer in decoder with bilinear upsampling layer and convolution layer.

Simply relying on blurred color domain causes difficulty to generate details. Motivated by image inpainting \cite{nazeri2019edgeconnect,song2018spg,dupont2018probabilistic}, we take advantage of the masked ground truth image to force generator into learning the details of the covered part. In the meantime, the input of edge is taken into more consideration by the network. Assuming that $X_{gt}$ is the ground truth image, $M$ is the binary random mask which will not cover more than 70\% area, and $E$ is the edge extracted from $X_{gt}$ as we discussed in Section \ref{Preprocessing of training Data}. We denote the output in the \emph{Imitation} \emph{Phase} as $X_{fake\text{-}1}$. We hope the output distribution $p(X_{fake\text{-}1})$ can be approximate as the distribution of ground truth image $p(X_{gt})$ when optimality is reached in the current phase.
\begin{eqnarray}
\label{G1-1}
&&X_{fake\text{-}1} = G_{1\text{-}1}(E , M \odot X_{gt})\\
&&p(X_{fake\text{-}1}) \Longrightarrow p(X_{gt})
\end{eqnarray}

The $G_{1\text{-}2}$ , our primary \emph{Generating} \emph{Phase}, continues to train after $G_{1\text{-}1}$ has converged. Since the generator has learned initialized features well, it enables generating more details and converges faster when the inputs are edge $E$ and color domain $C_{gt}$.

\begin{eqnarray}
\label{G1-2}
&&X_{fake\text{-}2} = G_{1\text{-}2}(E , C_{gt})\\
&&p(X_{fake\text{-}2}) \Longrightarrow p(X_{gt})
\end{eqnarray}
where $X_{fake\text{-}2}$ is the output in the \emph{Generating} \emph{Phase}.

The $G_{1\text{-}3}$ is the \emph{Refinement} \emph{Phase} inspired by Nazeri \emph{et} \emph{al}. \cite{nazeri2019edgeconnect}, which can reduce \emph{checkerboard} \emph{artifact} to generate more high frequency details and optimize the color distribution. $X_{fake\text{-}3}$ is the final output result.
\begin{eqnarray}
\label{G1-3}
&&X_{fake\text{-}3} = G_{1\text{-}3}(E , X_{fake\text{-}2})\\
&&p(X_{fake\text{-}3}) \Longrightarrow p(X_{gt})
\end{eqnarray}

{\bf Discriminator.}
$D_{1\text{-}1}$, $D_{1\text{-}2}$ and $D_{1\text{-}3}$ represent the three training phases of discriminator $D_1$ in turn. Just like $G_1$, there is only one discriminator $D_1$ all the time. We use PatchGAN \cite{isola2017image,yu2018free} architecture with spectral normalization \cite{miyato2018spectral} in the discriminator, which allows a larger receptive field to detect the generated fakes. Leaky ReLU activation function \cite{maas2013rectifier} rectifier is employed after each layers except for the last layer, where we use a sigmoid activation for the final output.
\begin{figure*}[t]
\centering


\includegraphics[width=0.9\linewidth]{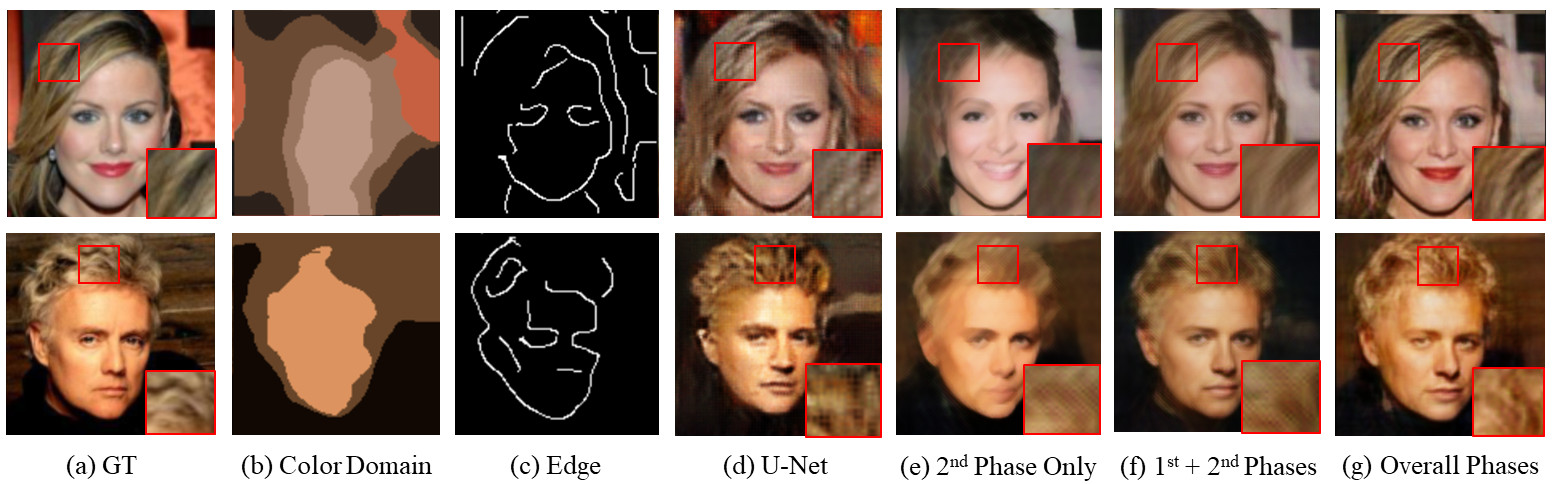}

\caption{\textbf{Output results compared among different generator architectures}: IR with U-net \cite{ronneberger2015u} from BicycleGAN \cite{zhu2017toward}, sole \emph{Generating} \emph{Phase} of PI-REC, \emph{Generating} \emph{Phase} with \emph{Imitation} \emph{Phase} and the overall phases.}
\label{Experiment.1}
\end{figure*}

\subsection{Model loss} \label{Model loss}
In $G_1$, we use a joint loss which contains per-pixel L1 loss, adversarial loss, feature loss \cite{johnson2016perceptual}, and style loss \cite{gatys2016image}. The overall
loss is calculated as below:
\begin{align}
\label{overall-loss}
L_{G_1} = &\alpha L_{per\text{-}pixel} + \beta L_{GAN\text{-}G} \\
          &+ \gamma L_{feature} + \delta L_{style}\notag\\
&L_{D_1} = L_{GAN\text{-}D}
\end{align}

{\bf Per-pixel Loss.}
Per-pixel loss is the $L_{1}$ loss difference between $X_{fake}$ and $X_{gt}$.
\begin{equation}
\label{Per-pixel Loss}
L_{per\text{-}pixel} = \frac{F_{sum}(X_{gt})}{F_{sum}(M)}\Arrowvert X_{fake} - M \odot X_{gt}\Arrowvert _{1}
\end{equation}
where function $F_{sum}()$ refers to the total number of non-zero pixels in the image. In $G_{1\text{-}1}$, if the mask has more covering area, the weight will be larger. In $G_{1\text{-}2}$ and $G_{1\text{-}3}$, the mask values are all non-zero so the weight remains the same value of one.

{\bf Adversarial loss.}
We choose LSGAN \cite{mao2017least} in order to create a stable generator which could fit the distribution of real images with high frequency details while traditional methods cannot.
\begin{align}
\label{Adversarial Loss}
L_{GAN\text{-}D} &= \frac{1}{2} \mathbb{E}[(D_1(X_{gt})-1)^2] \\
&+ \frac{1}{2}\mathbb{E}[D_1(G_1(E,I))^2]\notag
\end{align}
\begin{equation}
L_{GAN\text{-}G} = \frac{1}{2} \mathbb{E}[(D_1(G_1(E,I))-1)^2]
\end{equation}
where $G_1(E,I)$ represents the $X_{fake}$, and $I$ represents the different image input of each phases.

{\bf Feature loss.}
Feature reconstruction loss is included in the perceptual losses \cite{johnson2016perceptual}. Both low-level and high-level features are extracted from diverse convolutional layers in the pre-trained VGG19 network \cite{simonyan2014very} on the ImageNet dataset \cite{russakovsky2015imagenet}, which guarantees the perceptual content's consistency with the generated image.
\begin{equation}
L_{feature} =  \mathbb{E}[\sum_{i=1}^{L} \frac{1}{N_i}\Arrowvert(\Phi_i(X_{gt})-\Phi_i(X_{fake})\Arrowvert _{1}]
\end{equation}
where ${N_i}$ denotes the size of the $i\text{-}th$ feature layer and  $\Phi_i$ is the feature map of the $i\text{-}th$ convolution layer in VGG-19 \cite{simonyan2014very}. We use the feature map from layer $conv1\text{-}1$, $conv2\text{-}1$, $conv3\text{-}1$, $conv4\text{-}1$, $conv5\text{-}1$ other than using ReLU \cite{nair2010rectified} activation feature maps, which is aimed at generating sharper boundary lines suggested by ESRGAN \cite{wang2018esrgan}.

{\bf Style loss.}
Style reconstruction loss can also be included into perceptual losses \cite{johnson2016perceptual} which penalizes the differences in style.
\begin{equation}
L_{style} =  \mathbb{E}[\Arrowvert(G_{i}^{\Phi}(X_{gt})-G_{i}^{\Phi}(X_{fake})\Arrowvert _{1}]
\end{equation}
where  $G_{i}^{\Phi}$ is a Gram matrix of the $i\text{-}th$ feature layer. In addition, we find that the style loss can combat  the \emph{checkerboard} \emph{artifact} \cite{odena2016deconvolution} problem during \emph{Imitation} \emph{Phase}, while it barely works on other phases.

Note that we modify the hyperparameters values during different phases in order to get the desirable results. Specifically, in $Imitation$ $Phase$, we adopt $\alpha =1$, $\beta =0.01$, $\gamma = 1$ and $\delta = 150$. In the remaining phases, we increase the value of $\beta$ progressively to generate more high frequency details through generative adversarial loss. In the $2^{nd}$ phase $\beta$ is 0.1 and in the $3^{rd}$ $\beta$ turns into 2. $\delta$ is set to 0 in both latter two phases.

\begin{figure*}[t]
\centering


\includegraphics[width=0.95\linewidth]{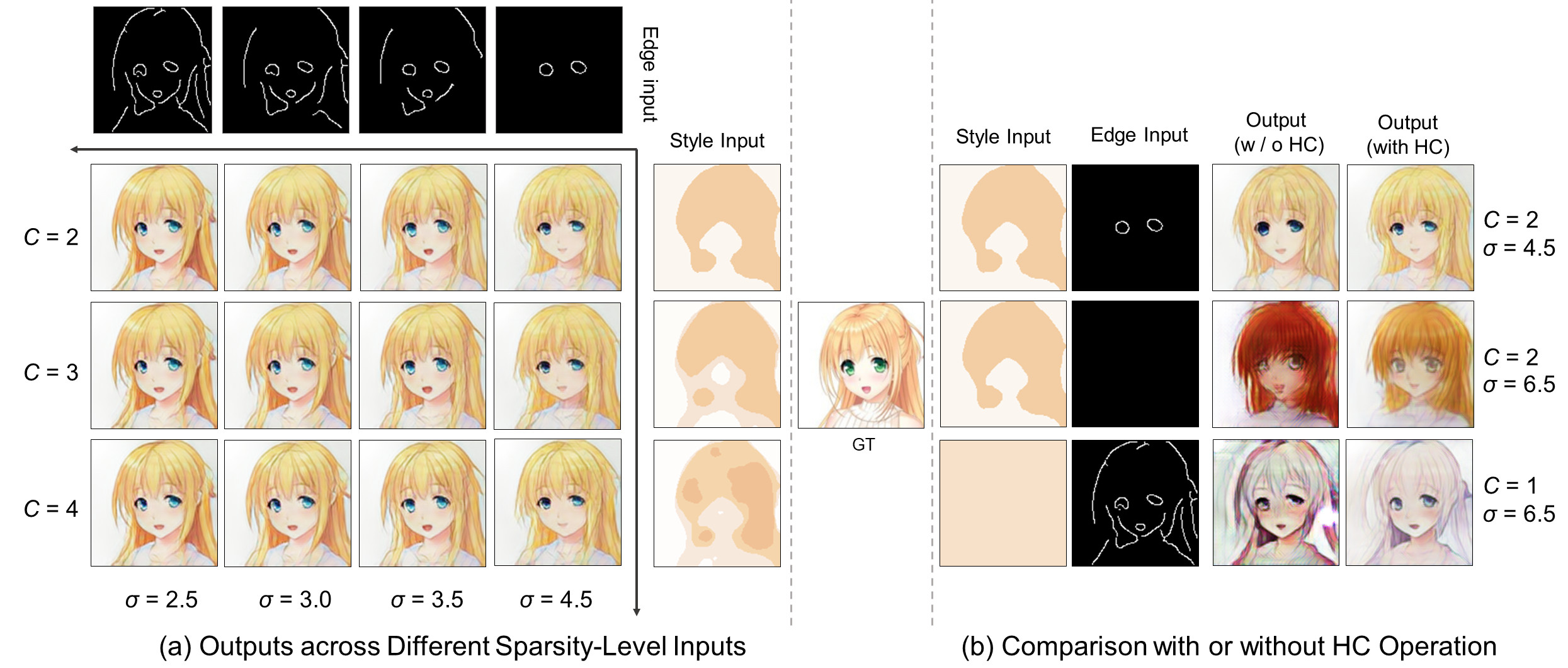}

\caption{(a) \textbf{Results across different sparsity-level inputs.} Owing to HC operation when training, we confirm that our model is not sensitive to certain fixed set of hyperparameters for testing. (b) \textbf{Comparison between output results with or without HC operation.} When employing HC operation in training, we can obtain better quality on local details and still get satisfactory outputs from extreme sparse content or style.}
\label{Experiment.2}
\end{figure*}

\section{Experiment}
\subsection{Datasets} \label{Datasets}
To train our model, we utilize dissimilar kinds of datasets: \emph{edges2shoes} \cite{isola2017image}, \emph{edges2handbags} \cite{isola2017image}, anime faces of \emph{getchu} \cite{jin2017towards} and \emph{CelebA} \cite{liu2015deep}. (Table \ref{dataset})

\begin{table}[ht]
\begin{tabular}{rcc}
\toprule
\multicolumn{1}{r}{Dataset} & Amount of Images & Size \\
\midrule
\textit{edges2shoes} & 50025 & 256x256 \\
\textit{edges2handbags} & 138767 & 256x256 \\
\textit{CelebA} & 203362 & 176x176 \\
\textit{getchu} & 34534 & 128x128 \\
\bottomrule
\end{tabular}
\caption{Information of datasets we adopt.}
\label{dataset}
\end{table}

\subsection{Ablation Study} \label{Ablation Study}
{\bf Advantage of Architecture.}
As we have discussed in Section \ref{Model Architecture}, our method has three progressive phases. As shown in Figure \ref{Experiment.1}, we demonstrate that our method has the advantage of reconstructing high frequency image. Specifically, we compare the U-net structure \cite{ronneberger2015u} with our $G_{1\text{-}2}$ architecture (Figure \ref{Experiment.1} (d, e)). U-net from BicycleGAN \cite{zhu2017toward} (pytorch version project) generates coarse high frequency details with more \emph{checkerboard}  \emph{artifact}, which causes great difficulty to improve quality pro\-gre\-ssive\-ly.

 \emph{Imitation} \emph{Phase} and \emph{Refinement} \emph{Phase} are also of appa\-rent significance in that they focus on generating high frequency details based on low frequency level, benefit from which the awful \emph{checkerboard} \emph{artifact} is almost eliminated (Figure \ref{Experiment.1} (e)). In addition, the color returns to a balanced level and more details of light and shadow are reproduced, as in ground truth images.

{\bf Sparsity of Inputs. }
As shown in Figure \ref{Experiment.2}, our model is not excessively sensitive to one fixed set of parameters,  where $C$ and $\sigma$ are the hyperparameters in K-means and Canny algorithm to control the sparsity. The outputs will be better if the inputs are more detailed as we expected.

As we have mentioned in Section \ref{Preprocessing of training Data}, hyperparameters confusion is another effective operation to ensure our model possessing a more powerful generalization ability to handle with inputs of various quality. Compared with fixing $C$ = 3 and $\sigma$ = 3 on training, the reconstruction outputs turn worse if the hyperparameters values are changed when testing (Figure \ref{Experiment.2} (b) top row), under which the refined details on eyes and hairs are lost. Furthermore, under the cases of extreme sparse inputs (Figure \ref{Experiment.2} (b) middle and bottom row), the outputs without HC operation is quite unsatisfactory. To sum up, with HC operation our model has a more powerful generalization and reconstruction ability.
%

\begin{figure*}[t]
\centering


\includegraphics[width=0.95\linewidth]{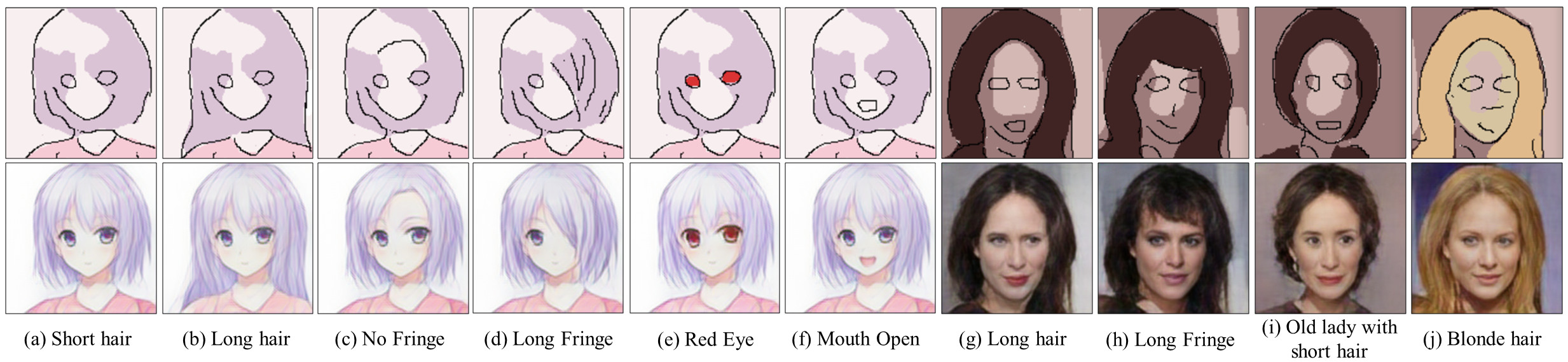}

\caption{\textbf{Hand drawn draft translation.} The top row denotes the hand draw drafts combined with edited edges and color domains. The bottom row illustrates the outputs, which are agilely responsive to the small changes in draft inputs.}
\label{Experiment.3}
\end{figure*}
\begin{figure*}[!t]
\centering
\includegraphics[width=0.95\linewidth]{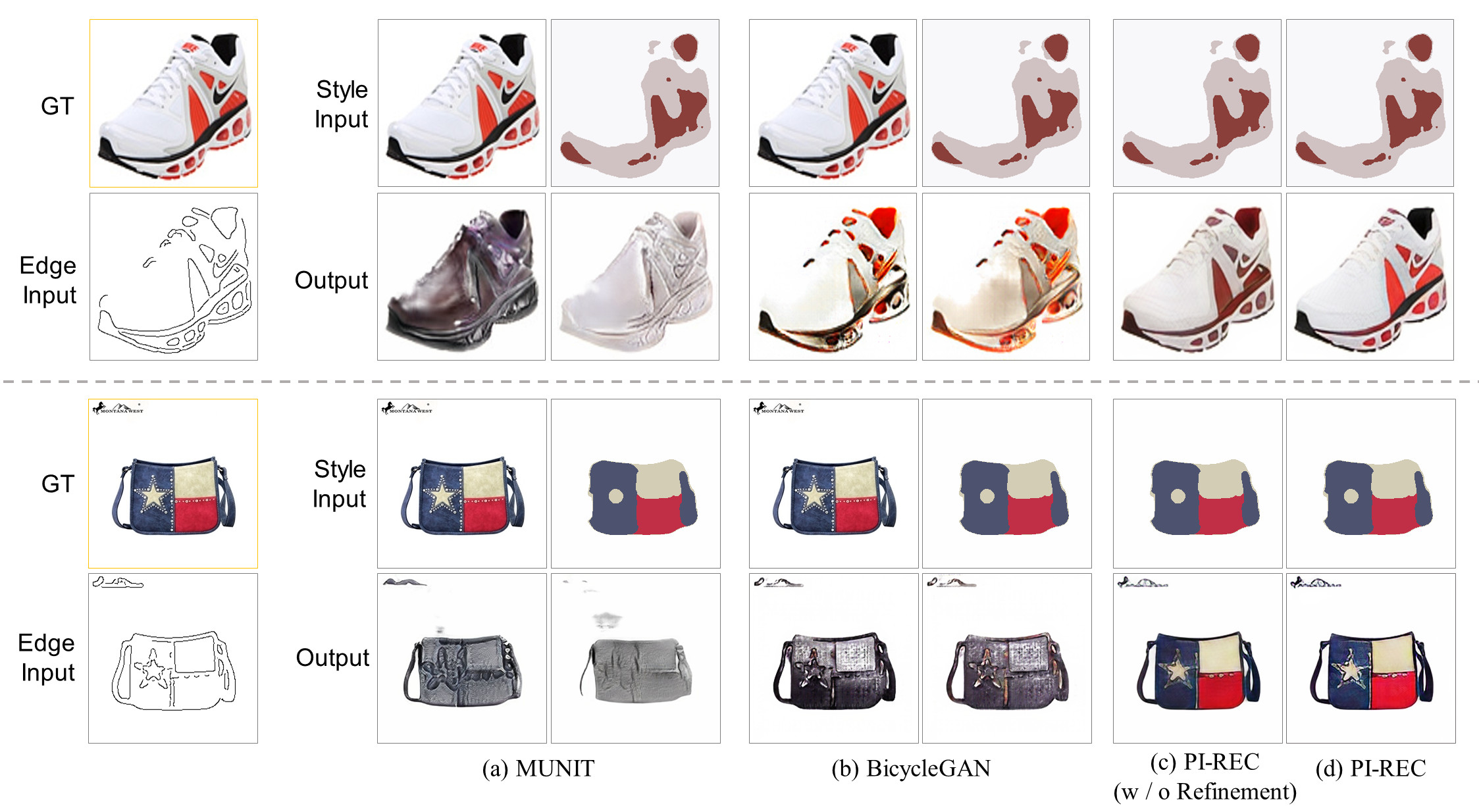}

\caption{\textbf{Qualitative compared results of PI-REC with baselines.} For MUNIT \cite{huang2018multimodal} and BicycleGAN \cite{zhu2017toward}, we use the ground truth image and color domain separately as style inputs, in order to obtain the best reconstruction outputs. Obviously, our model PI-REC with refinement can reconstruct the content and color details more accurately.}
\label{Experiment.4}
\end{figure*}

\begin{table*}[t]
\begin{center}

\setlength{\tabcolsep}{2.4mm}{
\begin{tabular}{lccccccccccc}
\toprule
\multicolumn{1}{c}{} & \multicolumn{5}{c}{edges $\rightarrow$ shoes} & \textit{} & \multicolumn{5}{c}{edges $\rightarrow$ handbags} \\  \cmidrule[0.4pt]{2-6}   \cmidrule[0.4pt]{8-12}
\multicolumn{1}{c}{} & \multicolumn{3}{c}{\textit{Realism}} &  & \textit{Accuracy} &  & \multicolumn{3}{c}{\textit{Realism}} & \textit{} & \textit{Accuracy} \\ \midrule[0pt]
\multicolumn{1}{c}{} & HP$^{*}$ & MMD & FID &  & LPIPS &  & HP$^{*}$ & MMD & FID &  & LPIPS \\ \cmidrule[0.2pt]{2-4} \cmidrule[0.2pt]{6-6}
\cmidrule[0.2pt]{8-10} \cmidrule[0.2pt]{12-12}
MUNIT$\rm_{gt}$ & - & 0.165 & 0.038 &  & 0.195 &  & - & 0.13 & 0.129 &  & 0.305 \\
MUNIT$\rm_{cd}$ & 12.50\% & 0.221 & 0.032 &  & 0.211 &  & 8.00\% & 0.195 & 0.083 &  & 0.336 \\
BicycleGAN$\rm_{gt}$ & - & 0.198 & 0.023 &  & 0.155 &  & - & 0.127 & 0.068 &  & 0.247 \\
BicycleGAN$\rm_{cd}$ & 33.00\% & 0.207 & 0.026 &  & 0.167 &  & 29.00\% & 0.145 & 0.074 &  & 0.253 \\
PI-REC$\rm_{w/o\_refine}$ & 44.20\% & \textbf{0.079} & 0.017 &  & 0.089 &  & 45.80\% & 0.118 & \textbf{0.067} &  & 0.171 \\
\textbf{PI-REC (ours)} & \textbf{62.30\%} & 0.081 & \textbf{0.015} &  & \textbf{0.085} &  & \textbf{57.10\%} & \textbf{0.112} & 0.069 &  & \textbf{0.168}
\\
\bottomrule
\end{tabular}}

\end{center}
\caption{\textbf{ Quantitative comparison results of PI-REC with baselines.}  \emph{cd} and \emph{gt} denote style inputs of color domain and ground truth respectively, \emph{w/o\_refine} denotes PI-REC without \emph{Refinement} \emph{Phase}. $^{*}$Higher is better while other metrics are opposite.}
\label{table:models2}
\end{table*}

\subsection{Qualitative Evaluation} \label{Qualitative Evaluation}
{\bf Hand Drawn Draft Translation.}
We design a painting software for drawing drafts, which records edges and color domain separately in turn. Moreover, we can see the real-time composite draft and outputs conveniently, as shown in  Figure \ref{Experiment.3} and Figure \ref{fig:banner} (b). The demo of this interactive software is shown in the supplementary material. For one thing, the edge plays an important role in generating content, which is not concrete but robust enough to generate various details like fringe (Figure \ref{Experiment.3} (c, d, h)), mouth (Figure \ref{Experiment.3} (f)) and hair (Figure \ref{Experiment.3} (a, b ,g)). For another thing, the flat color domain explicitly determines the global color distribution and gives ``hint" to local style specifically (Figure \ref{Experiment.3} (e, j)). In general, the model gets tradeoffs between edge and color domain for high-fidelity synthetics.


{\bf Comparison with Baselines.}
In Figure \ref{Experiment.4}, we qua\-li\-ta\-tively compare results of PI-REC with baselines on E2I tasks using \emph{edges2shoes} and \emph{edges2handbags} datasets. Our model outperforms state-of-the-art methods on both contents and style reconstruction. Regarding the content level, our model generates more accurate details (Figure \ref{Experiment.4} top half). On the style level, defects of using implicit style space occur if the input style is complex when there are two or more chief colors (Figure \ref{Experiment.4} bottom half). Despite using the ground truth as the style input, the extracted style vector with fixed length of eight commonly fails to contain enough information to represent image perfectly. Color distribution on details is thus lost and the rarely-exist color in datasets ends up being mapped into incorrect style vector space. Simply increasing the length of style vector or taking input of color domain as the style image also makes vain efforts on improving performance. On this point, without any strictly fixed style vector length, our model (Figure \ref{Experiment.4} (c, d)) with explicit style space can reconstruct the color details accurately.




\subsection{Quantitative evaluation} \label{Quantitative evaluation}
{\bf Evaluation Metrics.}
We evaluate the output results quantitatively on the aspects of realism and accuracy. For realism, we conduct human perceptual (HP) survey follow\-ed as Wang $et$ $al$. \cite{wang2018high}. Given pairs of generated images from various methods, five workers need to choose the more realistic one without time limit. Moreover, we use the kernel MMD \cite{gretton2007kernel} of the logits output and FID score \cite{heusel2017gans} to evaluate the output quality, which is recommended by Xu $et$ $al$. \cite{xu2018empirical}.

For evaluating reconstruction accuracy, we compute the average LPIPS distance \cite{zhang2018unreasonable} between ground truth image and reconstructed output in validation datasets. Lower scores (Equation \ref{Acc}) indicate that image pairs are more correlated based on human perceptual similarity.
\begin{equation}
\label{Acc}
Acc = \frac{1}{N} \sum \Phi_{LPIPS}(X_{gt}, G(E, S))
\end{equation}
where $N$ denotes the total number of sample pairs, and $G$ represents generator. $E$ and $S$ mean edge and style image respectively extracted from $X_{gt}$.

{\bf Realism  Accuracy Evaluation.}
As we depict in Table \ref{table:models2}, we compare our model with BicycleGAN \cite{zhu2017toward} and MUNIT \cite{huang2018multimodal}, which are the representative methods in supervised and unsupervised I2I translation domain respectively. We take input of the ground truth image to MUNIT and BicycleGAN as style image, in order to get the reconstruction result with best quality. In addition, for a fair comparison, we also input color domain to them as style image.

From the perspective of realism scores about MMD and FID, our model performs better than others as expected. The computed scores are close between PI-REC with refinement or not, since fine details generated by \emph{Refinement} \emph{Phase} is hard to catch by computed metric, while human can visually distinguish them.

With regard to reconstruction accuracy, lower LPIPS score is better according to Equation \ref{Acc}. Performance of MUNIT and BicycleGAN is nowhere near as accurate as PI-REC, the reason of which we have discussed in Section \ref{Qualitative Evaluation}.
\begin{figure}[h]
\centering


\includegraphics[width=1\linewidth]{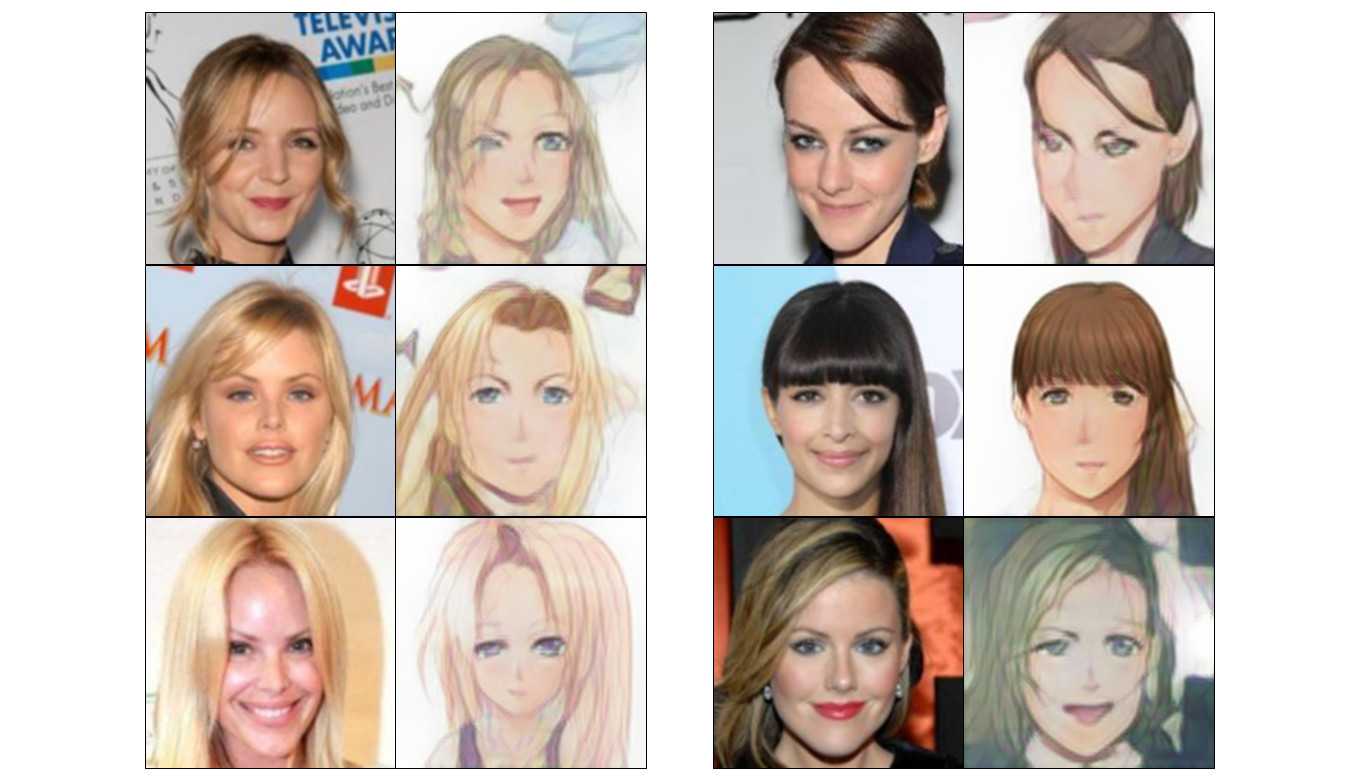}

\caption{\textbf{I2I translation with similar content.}}
\label{future}
\end{figure}

\section{Conclusion and Future Work}
We propose PI-REC, a novel progressive model for image reconstruction tasks. We achieve refined and high-quality reconstruction outputs when taking inputs of binary sparse edge and flat color domain only. The sparsity and interpretability of the inputs guarantee users with free and accurate control over the content or style of images, which is a significant improvement over existing works. Our method achieves state-of-the-art performance on standard benchmark of E2I task. Meanwhile, we obtain remarkable outputs in hand drawn draft translation tasks utilizing parameter confusion operation, which pushes the boundary of auto painting technology.

Our method can also be conditionally applied in I2I translation task if the contents between two domains are similar. As shown in Figure \ref{future}, we extract edge and color domain from realistic photos and feed them into well-trained model of anime. A few results are satisfactory on the texture of output paintings. We plan to combine the idea of cycle consistent loss into PI-REC to tackle with the user-defined style problem in the field of I2I translation.



{\small
\bibliographystyle{ieee}
\bibliography{egbib}
}

\newpage
\onecolumn
\appendix
\appendixpage
\addappheadtotoc

\begin{figure*}[h]
\centering


\includegraphics[width=0.8\linewidth]{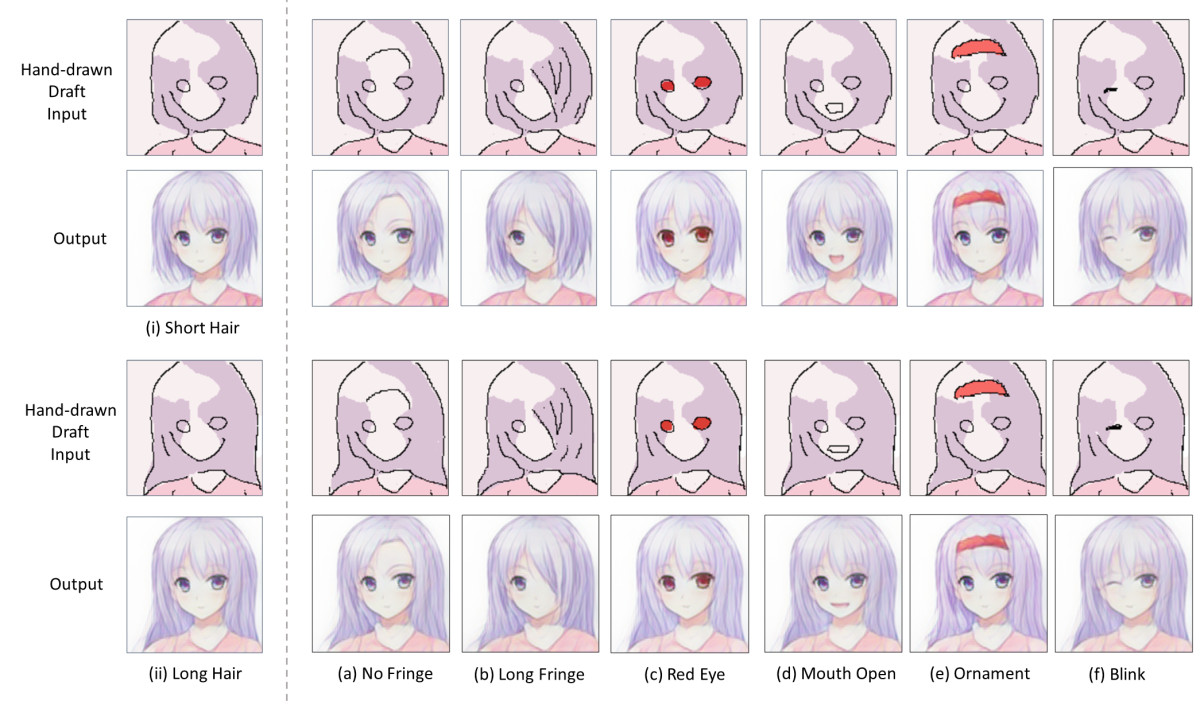}

\caption{\textbf{Hand drawn draft translation.} We conduct a controlled experiment on both (i) short hair and (ii) long hair conditions, which proves that our model has a strong compatibility.}

\label{future}
\end{figure*}

\begin{figure*}[h]
\centering
%
\includegraphics[width=0.75\linewidth]{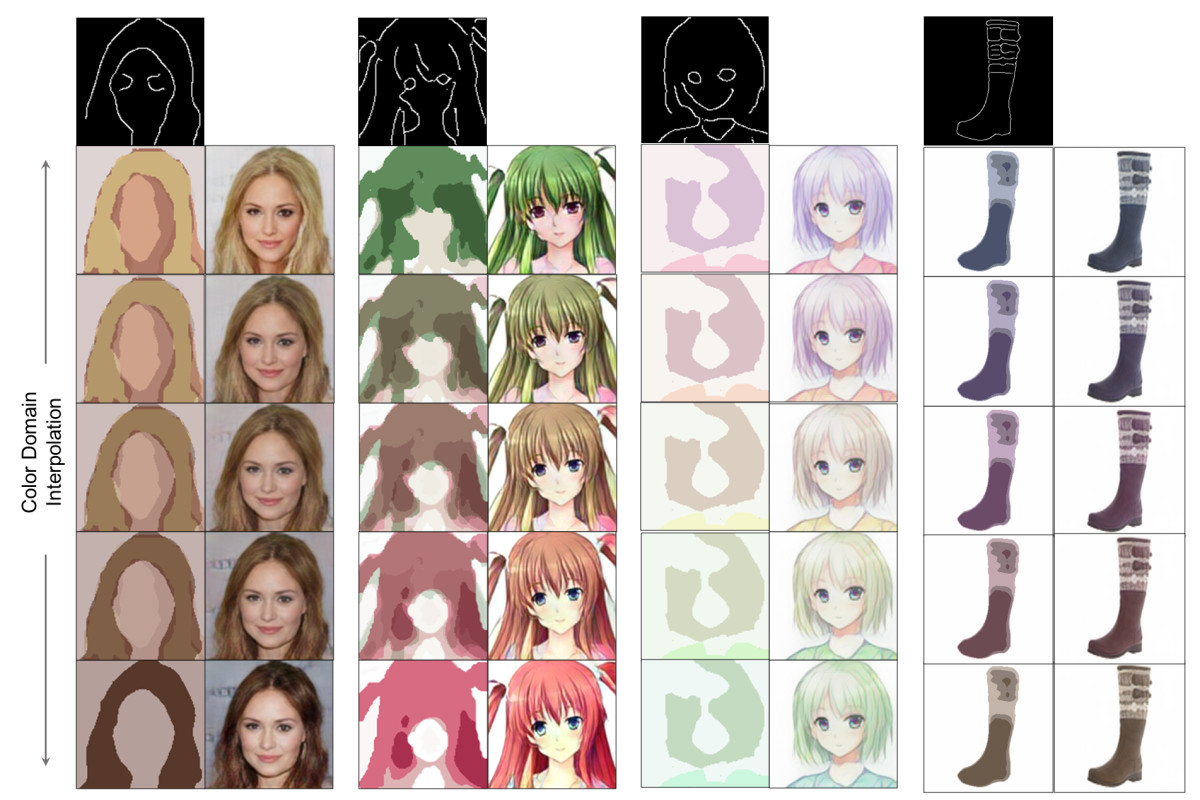}

\caption{\textbf{Interpolation between two color domains.} The same edge map and the interpolated color domain are taken as the input, which proves that our generator learns the color distribution from the explicit style space to generate the corresponding outputs.
}
\label{Experiment.5}
\end{figure*}

\begin{figure*}[t]
\centering


\includegraphics[width=1\linewidth]{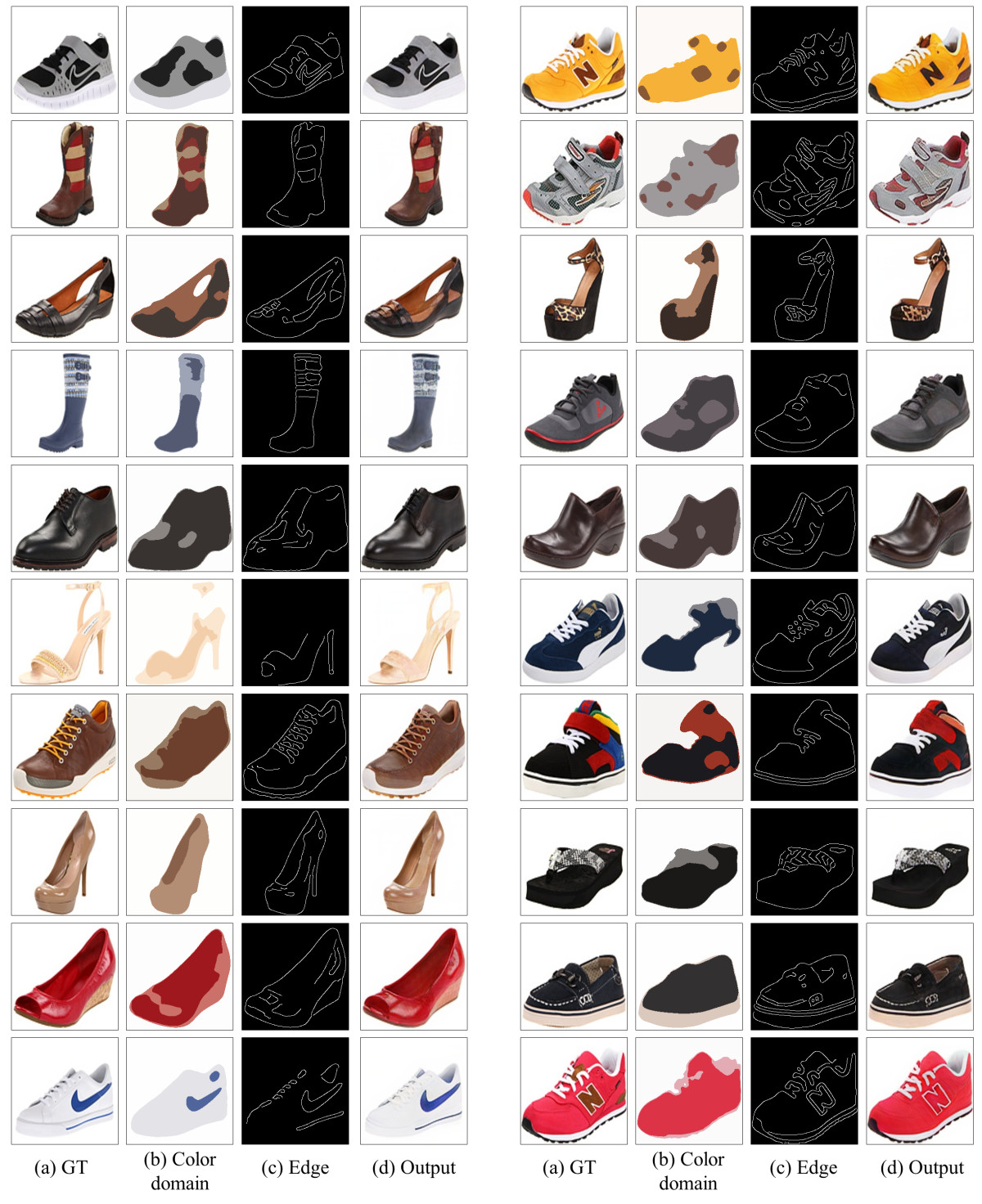}

\caption{\textbf{Image reconstruction on \emph{edges2shoes} dataset}}
\label{future}
\end{figure*}
\begin{figure*}[t]
\centering


\includegraphics[width=1\linewidth]{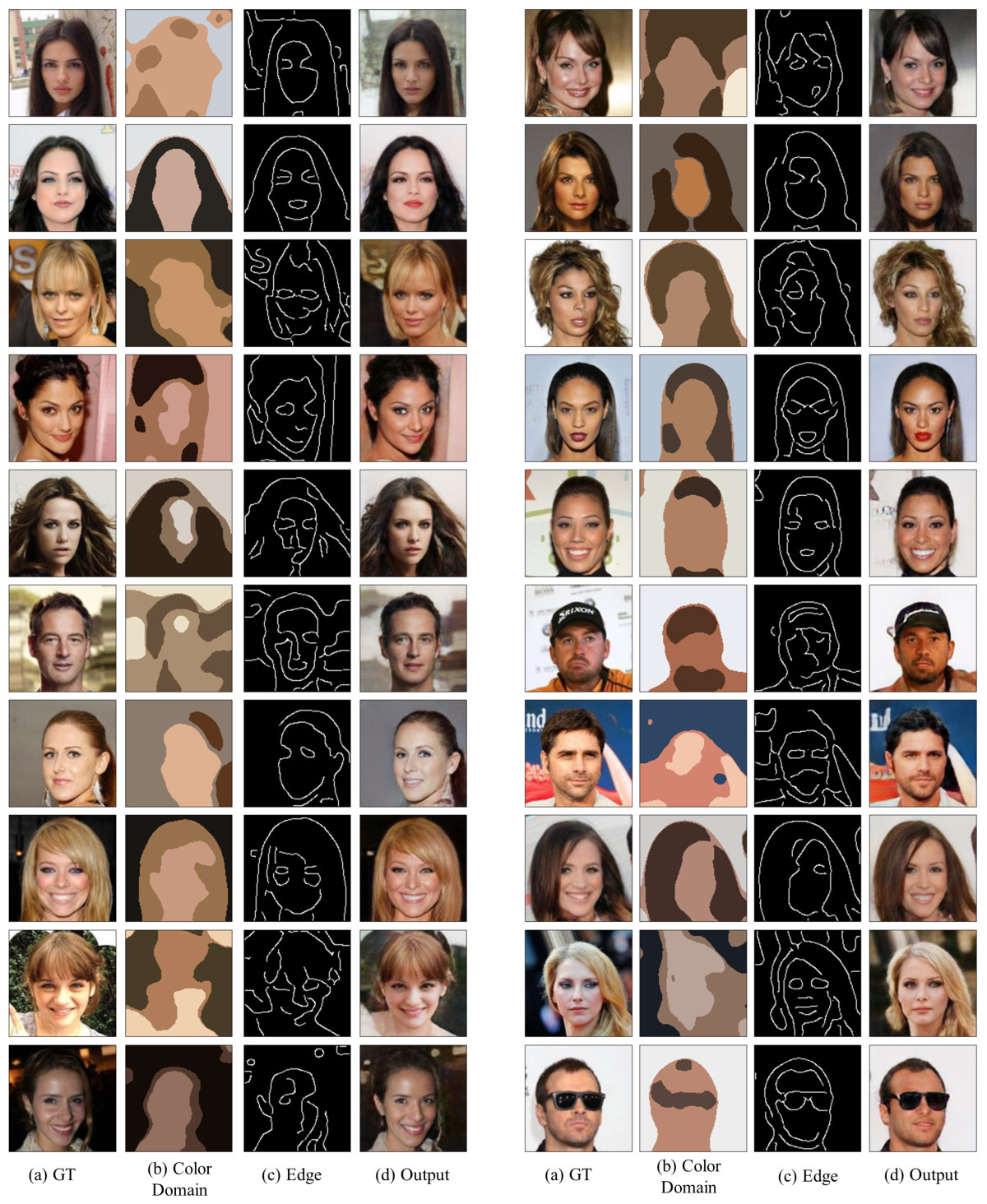}

\caption{\textbf{Image reconstruction on \emph{CelebA} dataset}}
\label{future}
\end{figure*}
\begin{figure*}[t]
\centering


\includegraphics[width=1\linewidth]{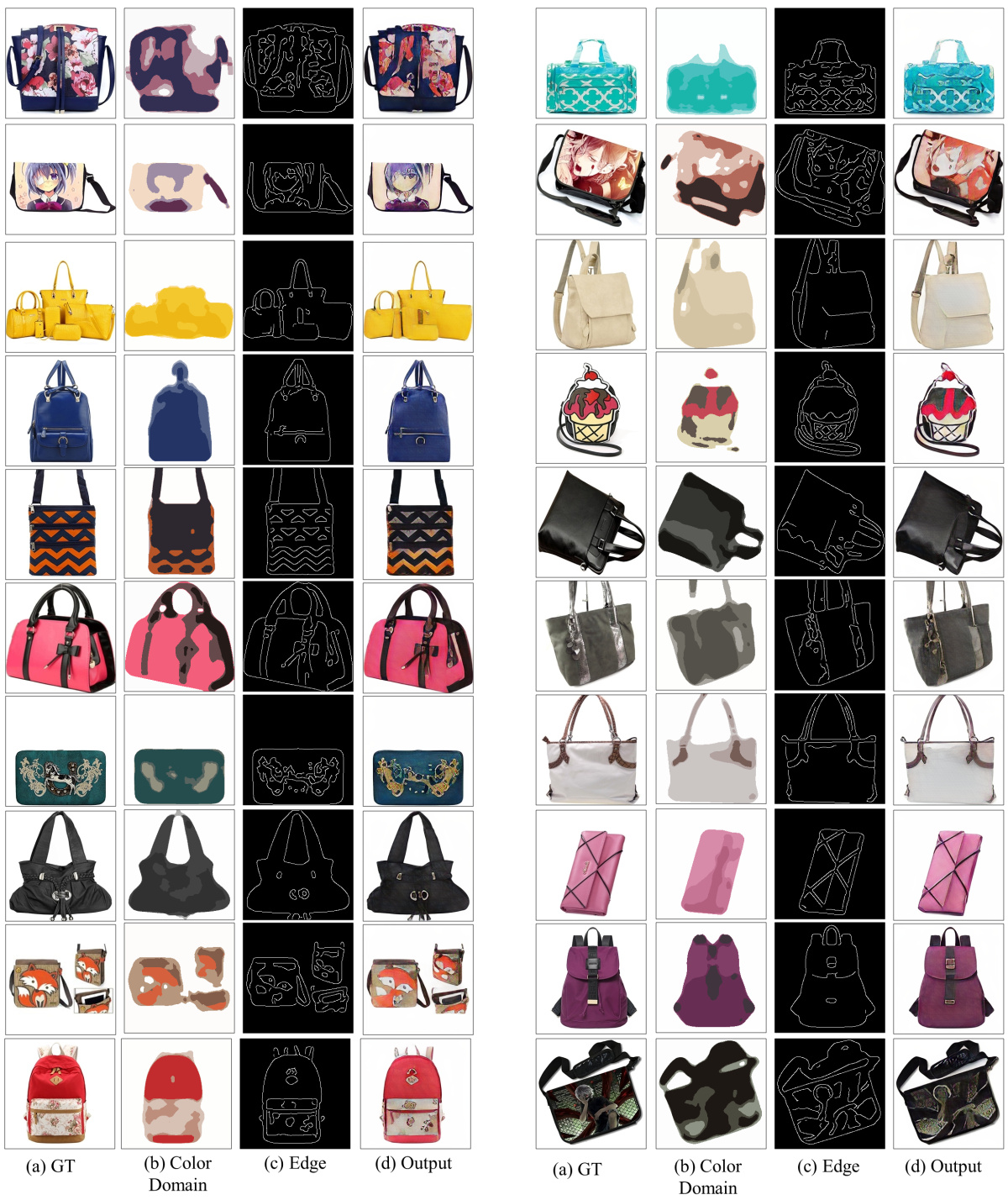}

\caption{\textbf{Image reconstruction on \emph{edges2handbags} dataset}}
\label{future}
\end{figure*}
\begin{figure*}[t]
\centering


\includegraphics[width=1\linewidth]{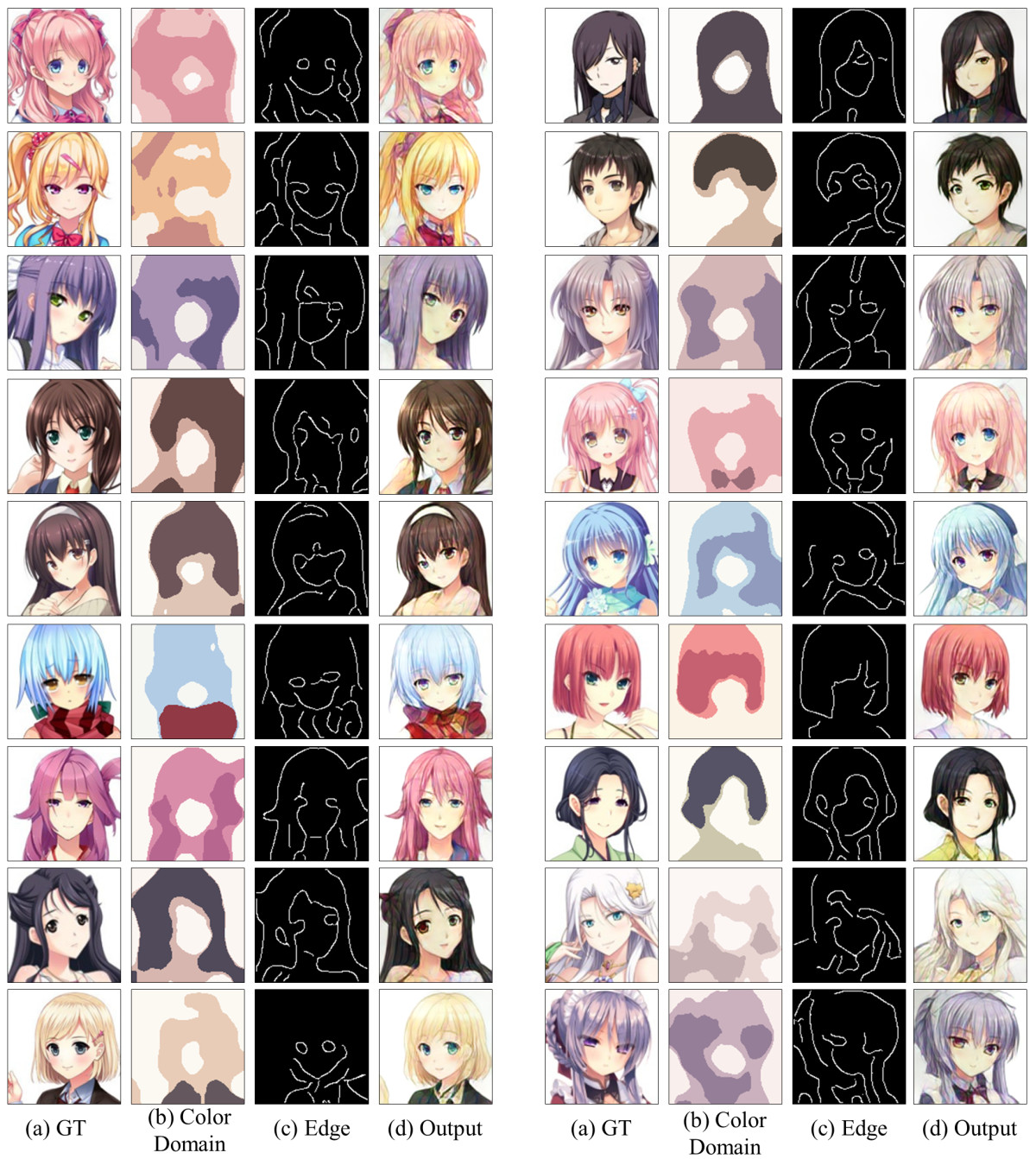}

\caption{\textbf{Image reconstruction on \emph{getchu} dataset}}
\label{future}
\end{figure*}

\end{document}